\documentclass[journal]{IEEEtran}

\usepackage[utf8]{inputenc}
\usepackage{graphicx}
\usepackage{amsmath}
\usepackage{amssymb}
\usepackage{acronym}
\usepackage[noadjust]{cite}
\usepackage{algorithm,algorithmic}
\usepackage{subfigure}

\newcommand{\FIG}[1]{Fig.~\ref{#1}}
\newcommand{\SEC}[1]{Section~\ref{#1}}
\newcommand{\TAB}[1]{Table~\ref{#1}}
\newcommand{\ALG}[1]{Algorithm~\ref{#1}}

\DeclareMathOperator*{\argmin}{argmin}

\acrodef{CNN}{convolutional neural network}
\acrodef{DR}{detection rate}
\acrodef{FDIA}{false data injection attack}
\acrodef{FedAvg}{federated averaging}
\acrodef{FedADAM}{federated ADAM}
\acrodef{FR}{false alarm rate}
\acrodef{GCN}{graph convolutional networks}
\acrodef{iid}[i.i.d.]{independent and identically distributed}
\acrodef{LSTM}{long short term memory}
\acrodef{MLP}{multi-layer perceptron}
\acrodef{PSSE}{power system state estimation}
\acrodef{ReLU}{rectified linear unit}
\acrodef{SGD}{stochastic gradient descent}
\acrodef{SVM}{support vector machine}
\acrodef{WLSE}{weighted least squares estimation}

\begin{document}
	
\title{Federated Learning Based Distributed Localization of False Data Injection Attacks on Smart Grids}

\author{Cihat~Ke\c{c}eci,
	Katherine~R.~Davis,~\IEEEmembership{Senior~Member,~IEEE},
	and~Erchin~Serpedin,~\IEEEmembership{Fellow,~IEEE}%
	\thanks{This work was supported by National Science Foundation under Grant 2220347. The statements made herein are solely the responsibility of the authors. (Corresponding author: Cihat Ke\c{c}eci.)}%
	\thanks{Cihat Ke\c{c}eci, Katherine R. Davis, and Erchin Serpedin are with the Department of Electrical and Computer Engineering, Texas A\&M University, College Station, TX 77843, USA (e-mail: \{kececi,katedavis,eserpedin\}@tamu.edu).}%
}

\maketitle

\begin{abstract}
Data analysis and monitoring on smart grids are jeopardized by attacks on cyber-physical systems.
False data injection attack (FDIA) is one of the classes of those attacks that target the smart measurement devices by injecting malicious data.
The employment of machine learning techniques in the detection and localization of FDIA is proven to provide effective results.
Training of such models requires centralized processing of sensitive user data that may not be plausible in a practical scenario.
By employing federated learning for the detection of FDIA attacks, it is possible to train a model for the detection and localization of the attacks while preserving the privacy of sensitive user data.
However, federated learning introduces new problems such as the personalization of the detectors in each node.
In this paper, we propose a federated learning-based scheme combined with a hybrid deep neural network architecture that exploits the local correlations between the connected power buses by employing graph neural networks as well as the temporal patterns in the data by using LSTM layers.
The proposed mechanism offers flexible and efficient training of an FDIA detector in a distributed setup while preserving the privacy of the clients.
We validate the proposed architecture by extensive simulations on the IEEE 57, 118, and 300 bus systems and real electricity load data.
\end{abstract}


\section{Introduction}

Smart grids incorporate information technology systems in sensing, processing, intelligence, and control of the power systems for robust transmission and distribution of electricity \cite{yu2016smart}.
While smart measurement devices and coupled communication networks bring many benefits and robustness to the power system, adversaries may alter the measurements and the other system parameters by attacking those nodes.
Cyberattacks on smart grids may cause significant problems in the operations and consequently, may interrupt the delivery of electricity in the system and result in economic losses.
The \acp{FDIA} pose a critical threat to the smart grids, and  are defined as the malicious activities carried out by an adversary via the injection of forged data into the information and control systems of smart grids.

Recently, machine learning-based \ac{FDIA} detection methods are emerging to efficiently detect the cyberattacks on smart grids. The complex rapidly-changing nature of smart grids makes it harder to efficiently and successfully detect the \acp{FDIA} using the conventional rule-based or deterministic approaches.
Machine learning algorithms enable efficient and robust detection and localization of the \acp{FDIA} by enabling the analysis of large volume data, identifying the complex hidden patterns from the historical data.

Recently, federated learning algorithms provide an efficient framework for training machine learning models on the edge devices.
The \ac{FedAvg} algorithm is proposed in \cite{mcmahan2017communication}. In \ac{FedAvg}, the local parameter updates of the local models are aggregated in the central server by taking the weighted average of the client parameter updates.
However, the \ac{FedAvg} algorithm performs poorly for the non-\ac{iid} data.
There have been attempts to address the problems in the federated learning environments.
The use of federated versions of the adaptive optimizers, such as ADAGRAD, ADAM, and YOGI is examined in \cite{reddi2021adaptive}. It was shown that using the adaptive optimizers speeds up the training of the federated deep learning models and increases the performance.
Federated learning has a wide range of applications such as analysis of the mobile user behavior, learning pedestrian patterns for autonomous vehicles, predicting and detecting health health-related events from wearable devices \cite{li2020federated}.

Federated learning is a promising solution for detecting the \acp{FDIA} on smart grids since it avails distributed training of the machine learning-based detectors. Federated learning solves the problem of cooperation between the electricity providers. Due to the confidentiality of the electricity power data, different electricity providers may not be willing to share their own data and it prevents collaborative training of a detection model. Federated learning enables the training of distributed local models without sharing the sensitive client data by sharing only the weights of the machine learning models.
Additionally, federated learning provides efficient training of the machine learning models by moving the computational burden from a centralized machine to distributed nodes.

\subsection{Related Work}

Comprehensive surveys on the problem of detection of \ac{FDIA} in smart grids are provided in \cite{deng2016false} and \cite{liang2016review}.
The name \ac{FDIA} was first used in \cite{liu2009false}.
A Kalman filter based \ac{FDIA} detector is proposed in \cite{manandhar2014detection}.
\ac{FDIA} is considered in \cite{guan2017distributed} using network theory.
The graph structure of the power grids is exploited for the detection of \acp{FDIA} in smart grids in \cite{drayer2020detection}.
An algorithm for detecting and compensating an \ac{FDIA} using a state estimation algorithm is proposed in \cite{khalaf2019joint}.
An \ac{FDIA} detection and localization algorithm utilizing interval observers that utilizes the bounds of the internal states, modeling errors, and disturbances is reported in \cite{luo2021interval}.
However, the performance of conventional methods is not sufficient (in terms of stability and accuracy) for the \ac{FDIA} detection problem in smart grids due to the complex and dynamic nature of the power grids as well as the sophisticated patterns in  cyberattacks.

Recently, the application of machine learning-based \ac{FDIA} detection methods has attracted a lot of interest in the literature. Machine learning algorithms provide a powerful toolset for the detection, localization, and identification of the \acp{FDIA}.
Batch and online learning algorithms are used to detect \ac{FDIA}s in \cite{ozay2015machine}.
An \ac{FDIA} detection algorithm that employs a  \ac{SVM} in combination with the alternating direction method of multipliers is proposed in \cite{esmalifalak2014detecting}. Study 
\cite{james2018online} exploits the wavelet transform and deep neural networks  to identify \ac{FDIA}s, while 
\cite{jevtic2018physics} employs a \ac{LSTM}  neural network to detect \ac{FDIA} attacks.
A \ac{CNN} model is used for \ac{FDIA} detection in \cite{wang2020locational}.
More recently, a graph neural network architecture was employed in \cite{boyaci2021graph, boyaci2021joint} to capture  the spatially localized features in the power grid for improved  detection of \ac{FDIA}s.

There are several studies in the literature that consider federated learning for smart grids. The energy demand for electric vehicle networks is predicted using federated learning in \cite{saputra2019energy}. Study \cite{taik2020electrical} proposes a federated learning method for forecasting household electricity load.
A federated learning-based approach for detecting \ac{FDIA} is proposed in \cite{li2022detection}. This study  proposes a transformer-based architecture for detecting \ac{FDIA}.
It uses the horizontal federated learning scheme in which all the clients must have the same data shapes. Hence, the approach proposed in \cite{li2022detection} trains the models at the bus level, which is not capable of utilizing the graph structure of the power network.
In this paper, we propose a novel hybrid deep neural network architecture consisting of graph neural network layers as well as \ac{LSTM} layers in a federated learning environment for  detection and  localization of \acp{FDIA}.
The proposed architecture is able to capture the hidden patterns in the power system data over both spatial and temporal dimensions. The \ac{GCN} layers are able to capture the local spatial correlations between the buses, while \ac{LSTM} helps to capture the temporal correlations. Our deep neural network architecture can be used on any partition of the power network. Therefore, it is generalizable and applicable to clients with any network structure, and it is not restricted to a single bus per client.

\subsection{Contributions}
The contributions of this work are next summarized. 
\begin{itemize}
	\item We propose a novel federated learning framework for distributed detection and localization of \ac{FDIA} attacks. The proposed scheme utilizes a two-layered architecture for the training of the attack detection model. The first group of layers are used as feature extractors and trained using the FedAvg algorithm. They consist of \ac{LSTM} layers, which are able to capture the time-domain patterns in the training data. On the other hand, the second group of layers consists of \ac{GCN} layers which capture the local spatial correlations between the power grid nodes. The second group of layers is trained using the FedGraph algorithm. This two-layered structure enables capturing the common time-domain patterns as well as the local spatial correlations between the nodes.
	\item We perform extensive simulations for the proposed scheme on a realistic simulation setup combining the IEEE bus systems and real power consumption data in comparison to the other machine learning-based detection methods. We performed the simulations for IEEE 57, 118, and 300 bus systems, respectively.
	The simulation results show that our method provides better F1-Score, detection, and false alarm rates compared to the other machine learning-based methods, such as federated transformer, \ac{LSTM}, and \ac{MLP}. The proposed scheme also provides a fair scheme for each client in the network since the distribution of the error metrics is uniform among each client.
\end{itemize}

The rest of the paper is organized as follows. A background on the \ac{FDIA} and federated learning is provided in \SEC{sec:background} The proposed method is introduced and discussed in \SEC{sec:proposed-method}. Extensive simulations are performed and the results are discussed in \SEC{sec:experiments}. Finally, the conclusions are drawn in \SEC{sec:conclusion}.

\section{Background}
\label{sec:background}

\subsection{False Data Injection Attack}
The noisy data samples from the measurement devices call for appropriate filtering, estimation and detection methods. Therefore, the operators of smart grids employ \ac{PSSE} in order to estimate the system state $x$  from the noisy samples $z$.
The nonlinear relationship between the system state and the measured noisy samples is defined by the measurement function $h(\cdot)$, which is expressed as:
\begin{equation*}
	z = h(x) + e,
\end{equation*}
where $e$ stands for the additive measurement noise.
The estimate $\hat{x}$ of the system state $x$ is found by
\begin{equation*}
	\hat{x} = \argmin_{x}  \left( z - h(x) \right) ^T R^{-1} \left( z - h(x) \right),
\end{equation*}
where $R$ represents the error covariance matrix.
The relationship between the active and reactive power injections at buses $P_i,Q_i$ and power flows between buses $P_{ij}, Q_{ij}$ are defined by the following equations \cite{abur2004power}: 
\begin{align*}
	P_i &= \sum_{j} V_i V_j \left( G_{ij} \cos\theta_{ij} + B_{ij} \sin\theta_{ij} \right) \\
	Q_i &= \sum_{j} V_i V_j \left( G_{ij} \sin\theta_{ij} - B_{ij} \cos\theta_{ij} \right) \\
	P_{ij} &= V_i^2 \left( g_{si} + g_{ij} \right) - V_i V_j \left( g_{ij} \cos\theta_{ij} + b_{ij} \sin\theta_{ij} \right) \\
	Q_{ij} &= -V_i^2 \left( b_{si} + b_{ij} \right) - V_i V_j \left( g_{ij} \sin\theta_{ij} + b_{ij} \cos\theta_{ij} \right).
\end{align*}
The provided system of equations can be solved using the \ac{WLSE} algorithm \cite{handschin1975bad}.
The bad data in the measurement data may adversely affect the system control and the performance in the power grids. In order to detect the bad data, the residual test is used in the power systems. The residual is defined as
\begin{equation*}
	\lVert r \rVert _2 = \lVert z - h(\hat{x}) \rVert _2.
\end{equation*}
If the residual is too large, it means that there is an error in the measurement values. Hence, the bad data can be detected by comparing the residual to a threshold value $\tau$:
\begin{equation*}
	\lVert r \rVert _2 \gtrless \tau.
\end{equation*}
\ac{FDIA} attacker tries to deviate the measurement values by injecting false data $a$ into the benign measurement values by forcing the solution of the power system state estimation algorithm to another point in the space. Consider that the benign measurement value is defined as
\begin{equation*}
	z_0 = h(x) .
\end{equation*}
The adversary  remains undetected by the system state estimation algorithm if  the added value $a$ to the measurement $z$ satisfies the condition:
\begin{equation*}
	a = h(\hat{x}+c) - h(\hat{x}).
\end{equation*}
This is due to the fact that the residual for the false data remains the same as the residual of the original data, i.e.,
\begin{align*}
	\lVert r_a \rVert _2 &= \lVert z_a - h(\hat{x}_a) \rVert _2 \\
	&= \lVert z_0 + a - h(\hat{x} + c) \rVert _2 \\
	&=  \lVert z_0 + h(\hat{x}+c) - h(\hat{x}) - h(\hat{x} + c) \rVert _2 \\
	&= \lVert r \rVert _2.
\end{align*}

\subsection{Federated Learning}
\label{sec:fed-learn}

Federated learning is a paradigm that enables joint training of a machine learning model using the distributed data from different client devices. Federated learning algorithms utilize the client devices for computing the local iterations and share those parameter updates with a central server for further processing and aggregation of the parameters.
In a typical federated learning setup, there are $M$ clients with $n_m$ local private data samples, which adds up to a total of $N=\sum_{m=1}^{M} n_m$ samples. The objective of the $m^{th}$ client is to minimize the given loss function
\begin{equation*}
	\min_{w} \frac{1}{n_m} \sum_{i=1}^{n_m} \ell_m \left( y_i, x_i; w \right),
\end{equation*}
where $x_i$, $y_i$, and $w$ are the features, labels, and weights of the machine learning model. Hence, the objective function at the server is expressed as 
\begin{equation*}
	\min_{w} \sum_{m=1}^{M} \frac{n_m}{N} \ell_m \left( y_m, x_m; w \right).
\end{equation*} 
In each global round of the federated learning algorithm, the server selects a subset $\mathcal{S}_t$ of the clients, where $|\mathcal{S}_t| << M$ and broadcasts the up-to-date model weights with those clients. Then, each client runs the \ac{SGD} algorithm using its own dataset for $E$ epochs. At the end of $E$ local epochs, the client uploads the gradient updates $\nabla w_t^m$ to the server. After receiving all the local updates, the server aggregates those local updates. There are different aggregation algorithms proposed in the literature. The \ac{FedAvg} \cite{mcmahan2017communication} algorithm aggregates the parameter updates by taking the weighted average of them as
\begin{equation*}
	\nabla w_t^{FedAvg} = \sum_{m=1}^{M} \frac{n_m}{N} \nabla w_t^m.
\end{equation*}
The FedAvg algorithm uses a fixed learning rate and the convergence rate may be slower for some problems.
In order to speed up the convergence rate of the machine learning model, \ac{FedADAM} \cite{reddi2021adaptive} algorithm aggregates and updates the local parameters using the ADAM \cite{kingma2014adam} optimizer. \ac{FedADAM} algorithm aggregates the parameters using the following equations
\begin{align*}
	\nabla w_t^{avg} &= \sum_{m=1}^{M} \frac{n_m}{N} \nabla w_t^m \\
	m_t &= m_{t-1} \beta_{1} + \left( 1 - \beta_{1} \right) \nabla w_t^{avg} \\
	v_t &= v_{t-1} \beta_{2} + \left( 1 - \beta_{2} \right) \left( \nabla w_t^{avg} \right)^2 \\
	\hat{m}_t &= \frac{m_t}{1 - \beta_{1}} \\
	\hat{v}_t &= \frac{v_t}{1 - \beta_{2}} \\
	w_t &= w_{t-1} - \eta \frac{\hat{m}_t}{\sqrt{\hat{v}_t} + \epsilon},
\end{align*}
where $\beta_{1}$, $\beta_{2}$, and $\epsilon$ are the hyperparameters. For instance, the values could be $\beta_{1}=0.9$, $\beta_{2}=0.999$, and $\epsilon=10^{-8}$.
As seen from the above equations, FedADAM adaptively adjusts the coefficients of the gradients, hence, provides a better convergence behavior for the deep neural network model.


\section{Proposed Method}
\label{sec:proposed-method}

\begin{figure}
	\centering
	\includegraphics[width=\linewidth]{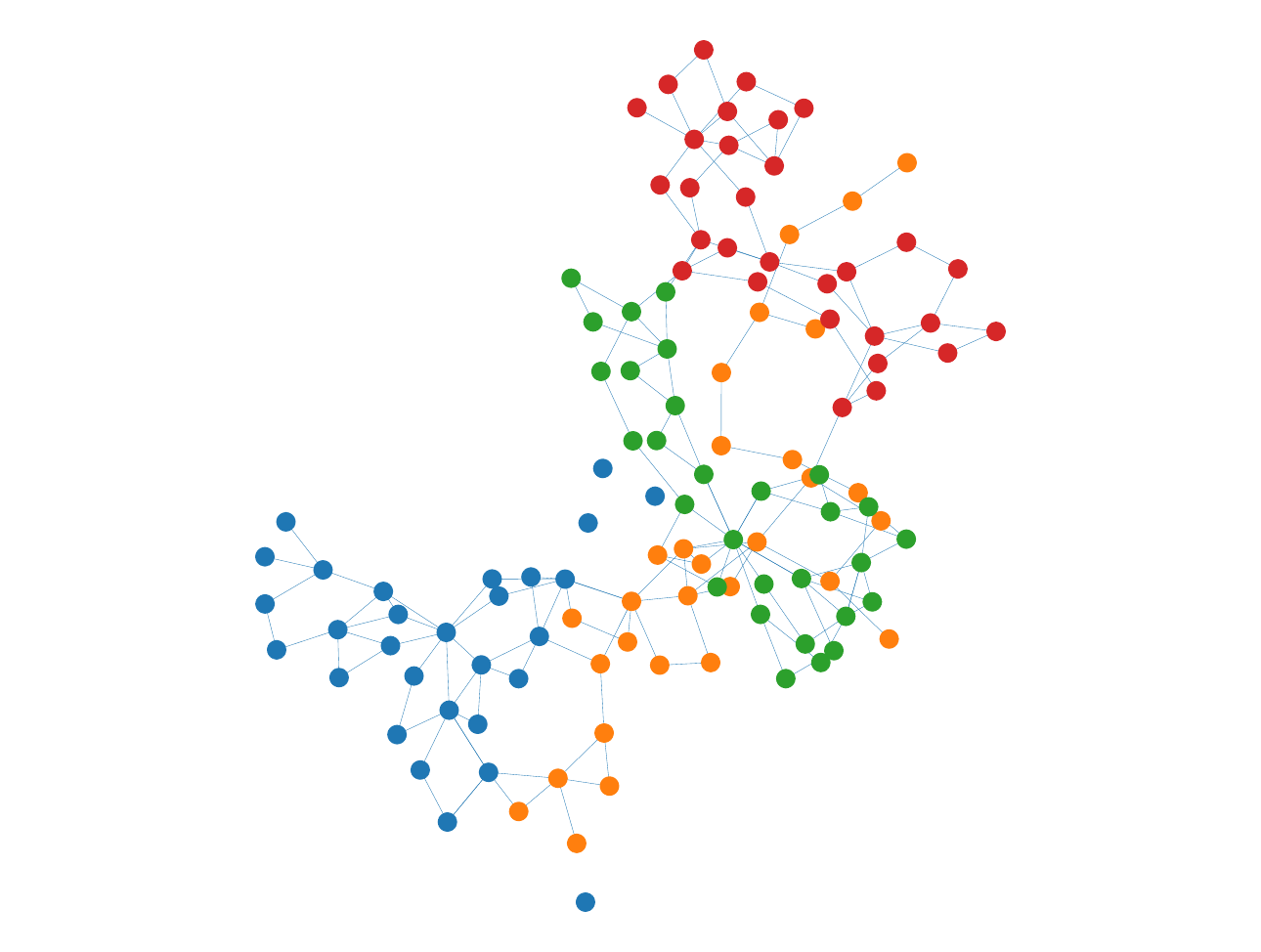}
	\caption{Federated learning enables training local models in each subnetwork cooperatively.}
	\label{fig:graph-partition}
\end{figure}

We consider a power network with multiple clients. Each client is responsible for a subnetwork in the power network. The clients have their own historical power load data for each bus in the respective subnetwork and are not willing to share it with the other clients. The aim of each client is to detect \acp{FDIA}. An example of partitioning of a power network graph is presented in \FIG{fig:graph-partition}. The buses belonging to the same client are shown with the same color.
We propose a federated learning-based method for training machine learning models for detection and localization of \acp{FDIA}.
By using a federated learning algorithm, we train a separate local model for each client and exchange only the required parameters between each client and a centralized coordination server.
We assume that there is a local machine at each client for computation of the model parameters and there is a connection between each client with the coordinating server.
As a deep neural network model, we use stacked \ac{LSTM} and \ac{GCN} layers in order to efficiently capture both the temporal and spatial patterns in the power data.

In order to capture the local spatial correlations in the power grid network, we represent the power grid as a graph. 
We denote the graph modeling the power grid by $\mathcal{G} = (\mathcal{V}, \mathcal{E})$, where $\mathcal{V}$ represents the set of vertices (nodes)  that corresponds to the set of buses in the power grid, and $\mathcal{E}$ stands for the set of edges that identify the power lines in the power grid.
The  graph adjacency matrix is denoted by $A \in \mathbb{R}^{|\mathcal{V}|\times |\mathcal{V}|}$, where $A_{i,j}=1$ if there is an edge between $i$ and $j$, $A_{i,j}=0$ otherwise. The degree matrix $D \in \mathbb{R}^{|\mathcal{V}|\times |\mathcal{V}|}$ is defined as the  diagonal matrix with main diagonal entries $D_{i,i} = \sum_{j} A_{i,j}$. The normalized Laplacian matrix is defined as $L =  I_n - D^{-\frac{1}{2}} A D^{-\frac{1}{2}}$, where $I_n$ denotes the identity matrix.
For each node, we have the feature vector $x(v) = \lbrace p_v, q_v \rbrace$ that incorporates the active and reactive powers. The decision variable indicating the event of an \ac{FDIA} attack is denoted by $d(v)$.

The detection and localization algorithm is deployed in a distributed manner for a more efficient evaluation and fast response to the attacks. The nodes are distributed to local servers by their physical locations. Each local server corresponds to the clients in the federated learning terminology. The set of all servers is denoted by $\mathcal{C}$. The subgraph and the Laplacian that belong to the client $c$ are denoted by $\mathcal{G}_c = (\mathcal{V}_c, \mathcal{E}_c)$, and $L_c$, respectively.

\begin{figure}
	\centering
	\includegraphics[width=0.7\linewidth]{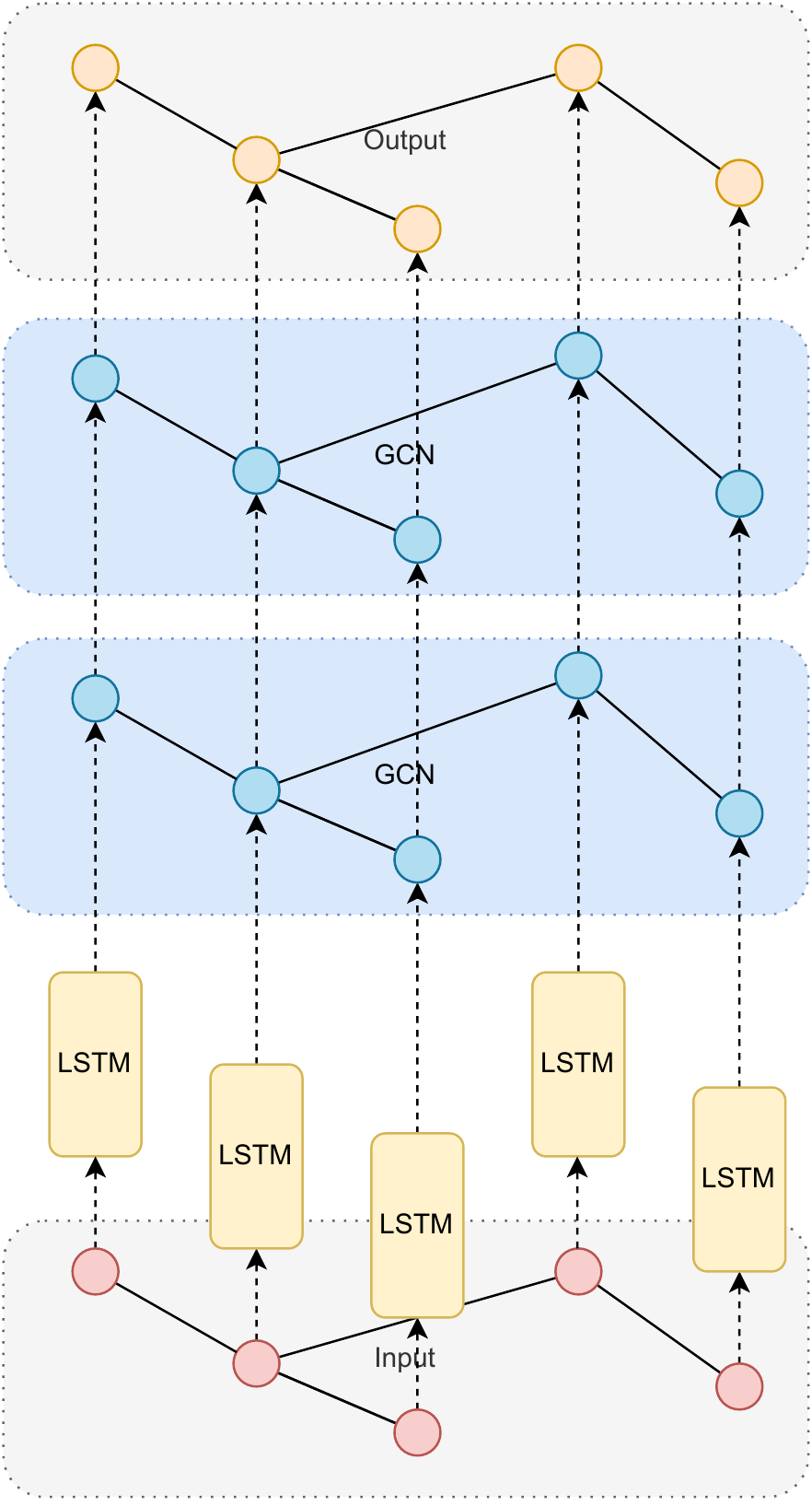}
	\caption{The proposed architecture for the machine learning model. The LSTM layers are used at the node level and the GCN layers are used at the graph level for each client in the network.}
	\label{fig:architecture}
\end{figure}

Two different groups of layers are used in the machine learning model. The proposed machine learning architecture is shown in \FIG{fig:architecture}.

The first layer group consists of \ac{LSTM} layers \cite{hochreiter1997lstm}  and they are used as feature extractors.
The input-output relationship for an \ac{LSTM} layer assumes the following dynamics. First, the forget gate unit $g_f(x)$ assumes the following input-output relationship:
\begin{equation*}
	F_t = \sigma\left(W^f \cdot [H_{t-1}, X_t] + B^f \right),
\end{equation*}
where $X_t$ is the input, $H_{t}$ is the hidden state, $W^f$ and $B^f$ are the weight and bias for the forget gate. The internal state of the \ac{LSTM} cell is updated as follows
\begin{equation*}
	S_t = F_t S_{t-1} + I_t \sigma\left(W^s \cdot [H_{t-1}, X_t] + B^s \right).
\end{equation*}
The input to gate unit $I_t$ is given by
\begin{equation*}
	I_t = \sigma\left(W^i \cdot [H_{t-1}, X_t] + B^i \right),
\end{equation*}
while the gate output  takes the form: 
\begin{equation*}
	O_t = \sigma\left(W^o \cdot [H_{t-1}, X_t] + B^o \right).
\end{equation*}
Finally, the hidden state $H_t$ is calculated as
\begin{equation*}
	H_t = O_t \cdot \tanh (S_t).
\end{equation*}
The feature extractor \ac{LSTM} layers share the same weight values for each node in the graph and are trained jointly using the FedAvg algorithm \cite{mcmahan2017communication}, which is discussed in detail in \SEC{sec:fed-learn}.
The weights of the feature extractor layers are denoted as $W_{\text{fe}}^{l}$.

The second group consists of \ac{GCN} layers which extract the local correlations between the nodes in the network. Those layers are trained using the FedGraph \cite{chen2021fedgraph} algorithm, as explained below. 
The \ac{GCN} layers exploit the graph structure of the power grid network.
The weights of the \ac{GCN} layers are denoted as $W_{\text{gcn}}^{l}$.
The aggregation operation in a \ac{GCN} layer is given by
\begin{equation*}
	Z_{gcn}^{l+1} = L H_{gcn}^{l} W_{gcn}^{l}.
\end{equation*}
The output of a \ac{GCN} layer is calculated as
\begin{equation*}
	H_{gcn}^{l+1} = \sigma \left( Z_{gcn}^{l+1} \right),
\end{equation*}
where $\sigma(\cdot )$ represents the nonlinear activation function, such as ReLU, tanh, or sigmoid.

The training procedure for client $s\in \mathcal{S}$ is as follows. First, the client downloads the up-to-date weights $W_{\text{gcn}}$ from the server. After initializing the local weights, the client calculates
\begin{gather*}
	Z_{gcn}^{l+1} = \sum_{u \in \mathcal{V}_c} L_{c}^{l}(v,u) H_{gcn,c}^{l}(u) W_{gcn,c}^{l} +\\
	\sum_{c' \in \mathcal{C}, c\neq c'} \sum_{u \in \mathcal{V}_{c'}} L_{c}^{l}(v,u) H_{gcn,c'}^{l}(u) W_{gcn,c'}^{l}.
\end{gather*}
Then, the output is calculated as
\begin{equation*}
	H_{gcn,c}^{l+1}(v) = \sigma \left( Z_{gcn}^{l+1}(v) \right).
\end{equation*}
The clients only share the hidden embeddings with the other clients in the federated learning scheme. In this way, clients are able to keep their data private.

The overall training algorithm is provided in \ALG{alg:training}.

\begin{algorithm}
	\caption{Training procedure}
	\label{alg:training}
	\begin{algorithmic}
		\FOR{iteration $t=1$ {\bfseries to} $T$}
			\STATE broadcast the up-to-date weights to all clients $c\in \mathcal{C}$
			\FOR{each client $c \in \mathcal{C}$ {\bfseries in parallel}}
				\FOR{each node $v \in \mathcal{V}_c$ }
					\FOR{each feature extractor layer $l$}
						\STATE initialize the weights using up-to-date weights $W_{\text{fe,c}}^{l} = W_{\text{fe}}^{l}$
						\STATE compute the gradient $\nabla \ell \left( W_{\text{fe,c}}^{l} \right)$
						\STATE update local weights $W_{\text{fe,c}}^{l} = W_{\text{fe,c}}^{l} - \eta \nabla \ell \left( W_{\text{fe,c}}^{l} \right) $
					\ENDFOR
					\FOR{each GCN layer $l$}
						\STATE $Z_{gcn}^{l+1} = \sum_{u \in \mathcal{V}_c} L_{c}^{l}(v,u) H_{gcn,c}^{l}(u) W_{gcn,c}^{l} + \sum_{c' \in \mathcal{C}, c\neq c'} \sum_{u \in \mathcal{V}_{c'}} L_{c}^{l}(v,u) H_{gcn,c'}^{l}(u) W_{gcn,c'}^{l}$
						\STATE Generate embeddings $H_{gcn,c}^{l+1}(v) = \sigma \left( Z_{gcn}^{l+1}(v) \right)$
						\STATE compute the gradient $\nabla \ell \left( W_{\text{gcn,c}}^{l} \right)$
						\STATE update local weights $W_{\text{gcn,c}}^{l} = W_{\text{gcn,c}}^{l} - \eta \nabla \ell \left( W_{\text{gcn,c}}^{l} \right) $
					\ENDFOR
				\STATE broadcast $W_{\text{fe,c}}^{l}$ and $W_{\text{gcn,c}}^{l}$ back to server
				\ENDFOR
			\ENDFOR
			\STATE aggregate the weights for feature extractor layers
			\STATE $\nabla W_{\text{fe}}^{\text{avg}} = \sum_{m=1}^{M} \frac{n_c}{\sum_{c} n_c} \nabla W_{\text{fe,c}}$
			\STATE $m_t = m_{t-1} \beta_{1} + \left( 1 - \beta_{1} \right) \nabla W_{\text{fe}}^{\text{avg}}$
			\STATE $v_t = v_{t-1} \beta_{2} + \left( 1 - \beta_{2} \right) \left( \nabla W_{\text{fe}}^{\text{avg}} \right)^2$
			\STATE $\hat{m}_t = \frac{m_t}{1 - \beta_{1}}$
			\STATE $\hat{v}_t = \frac{v_t}{1 - \beta_{2}}$
			\STATE $\nabla W_{\text{fe}}^{(t)} = \nabla W_{\text{fe}}^{(t-1)} - \eta \frac{\hat{m}_t}{\sqrt{\hat{v}_t} + \epsilon}$
		\ENDFOR
	\end{algorithmic}
\end{algorithm}

\section{Experiments}
\label{sec:experiments}

In this section, we evaluate the performance of the proposed algorithm. We use the IEEE 57, 118, and 300 bus systems, respectively for modeling the power network structure and real hourly electricity load data for modeling the load in the network.

\subsection{Dataset Generation}
\begin{figure}
	\centering
	\includegraphics[width=\linewidth]{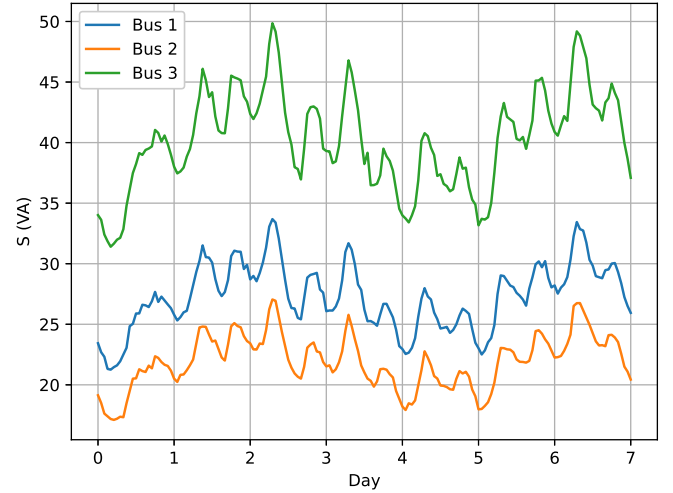}
	\caption{An example of the load profile generated for buses 1, 2, 3.}
	\label{fig:load-profiles}
\end{figure}

We used the IEEE 57, 118, and 300 bus systems, respectively for modeling the power network in the experiments. In order to model the power load profiles, we used the hourly power load dataset provided by ERCOT \cite{ercot2023}. We scaled the power load values in order to adapt to the parameters of the test grid we employ. For each bus in the network, we first scaled the mean value of the hourly load values to the base load of the bus. Then, we scaled the standard deviation of the load values to $0.01$ times the base load for the respective bus. A visualization of the generated measurement samples is provided in \FIG{fig:load-profiles} for buses $1$, $2$, and $3$.

We generated the malicious data samples using the following approach. First, we generated the benign data samples using the steps above. Then, an attacker selects a subset of the buses in the network. Then, performs an attack at the selected buses by one of the following methods: (i) random attack, the measurement vector is replaced by $z_t^a = z_t + \alpha_t^r$ \cite{boyaci2021joint}, where $\alpha_t^r$ is a uniform random variable $\alpha_t^r \sim \mathcal{U}(a, b)$ \cite{jevtic2018physics}, or a Gaussian random variable $\alpha_t^r \sim \mathcal{N}(1,\; \sigma_{z})$ \cite{ozay2015machine}; (ii) replay attack, the measurement vector is replaced by a previous value $z_t^a = z_{t-\delta}$ \cite{chaojun2015detecting}; (iii) scale attack, measurement vector is scaled by $z_t^a = \alpha_s z_{t}$ \cite{hasnat2020detection}.

We generated $8752$ data samples, which are split into training and test datasets with the ratio $80\%-20\%$. The attacked measurement values are uniformly selected from each attack type. 

\subsection{Hyperparameters}
We used a hybrid model as a deep learning architecture as discussed in the previous sections. The feature extractor layers are composed of two \ac{LSTM} layers each with $32$ units. The \ac{GCN} layer group consists of three \ac{GCN} layers with $128$, $128$, and $1$ output nodes per client, respectively. We used \ac{ReLU} activation function in each layer of the network. We used ADAM optimizer in the server and \ac{SGD} optimizer in the clients. We performed $1$ local epoch per training round. The training batch size is $64$. The local and global learning rate values are optimized using a grid search. We used the cross-entropy function as a loss function.
We used the Python programming language and Tensorflow \cite{tensorflow} package to conduct the simulation experiments.

\subsection{Benchmark Methods}
In order to benchmark the performance of the proposed method, we implemented and compared the results with the federated transformer, federated \ac{LSTM}, and federated \ac{MLP} algorithms.
The proposed federated graph learning algorithm can be trained on any partition of the graph even with different numbers of buses in each client while enabling the use of the \ac{GCN} architecture for training in a federated scheme.
However, the benchmark algorithms from the literature use the horizontal federated learning scheme in which the algorithm requires the shapes of the features of each client to be the same. Hence, this requirement limits the application of the federated learning algorithm at the node level. Therefore, the current methods assign a client for each bus in the network. For instance, for the IEEE 118 bus system, we have 118 clients in the network.
The federated transformer \cite{li2022detection} algorithm uses a transformer architecture as a local deep learning model for each client. Similarly, the federated \ac{LSTM} algorithm uses stacked \ac{LSTM} architecture for each client. As for the federated \ac{MLP} algorithm, we use a \ac{MLP} model for the client models.

\subsection{Results}
Since we have an unbalanced data distribution, i.e., the amount of benign data samples per bus is much higher than the malicious data samples, it is not plausible to use the accuracy values as an evaluation metric for better understanding. Hence, we use \ac{DR}, \ac{FR}, and F1-Score (F1) as evaluation metrics. The mentioned metrics are calculated as
\begin{align*}
	\text{DR} &= \frac{\text{TP}}{\text{TP} + \text{FN}} \\
	\text{FR} &= \frac{\text{FP}}{\text{FP} + \text{TN}} \\
	\text{F1} &= \frac{2\text{TP}}{2\text{TP}+\text{FP}+\text{FN}}.
\end{align*}
The corresponding values of F1, \ac{DR}, and \ac{FR} for each algorithm are shown in \TAB{tab:benchmark} and  are tested on the IEEE 57, 118, and 300 bus systems, respectively. As seen from the table, the federated graph learning algorithm outperforms the other algorithms by a significant margin.
The proposed algorithm is able to achieve F1-score values around $97-98\%$ for each IEEE test case. On the other hand, the second best performing algorithm, federated LSTM could only achieve an F1-score around $85-86\%$. The other methods fall below $80\%$ for each test case.
This is in line with the expected results because the federated graph learning algorithm is able to capture both the temporal and spatial patterns in the data. The proposed method performs well on each IEEE bus system.

\begin{table*}
	\centering
	\caption{Comparison of the F1-Score detection rate, and false alarm rate (in percentage) for each method.}
	\label{tab:benchmark}
	\begin{tabular}{r|ccc|ccc|ccc}
		\hline
		& \multicolumn{3}{c}{IEEE 57} & \multicolumn{3}{|c}{IEEE 118} & \multicolumn{3}{|c}{IEEE 300} \\
		\hline
		Method         &  F1   &  DR   &  FR   &  F1   &  DR   &  FR   &  F1   &  DR   &  FR  \\
		\hline
		FedGraph       & 97.90 & 96.03 & 0.01  & 97.86 & 95.92 & 0.00  & 98.46 & 96.99 & 0.00 \\
		FedLSTM        & 85.64 & 74.91 & 0.00  & 86.21 & 75.89 & 0.00  & 85.43 & 74.66 & 0.00 \\
		FedTransformer & 78.00 & 69.04 & 0.94  & 76.42 & 77.54 & 3.10  & 77.63 & 68.23 & 0.65 \\
		FedMLP         & 79.95 & 69.26 & 0.00  & 74.74 & 66.87 & 1.34  & 64.43 & 54.68 & 0.00 \\
		\hline
	\end{tabular}
\end{table*}

\begin{figure*}
	\centering
	\subfigure[FedGraph IEEE-57]{%
		\centering
		\label{fig:box-plot-fedgraph-57}%
		\includegraphics[width=.32\linewidth]{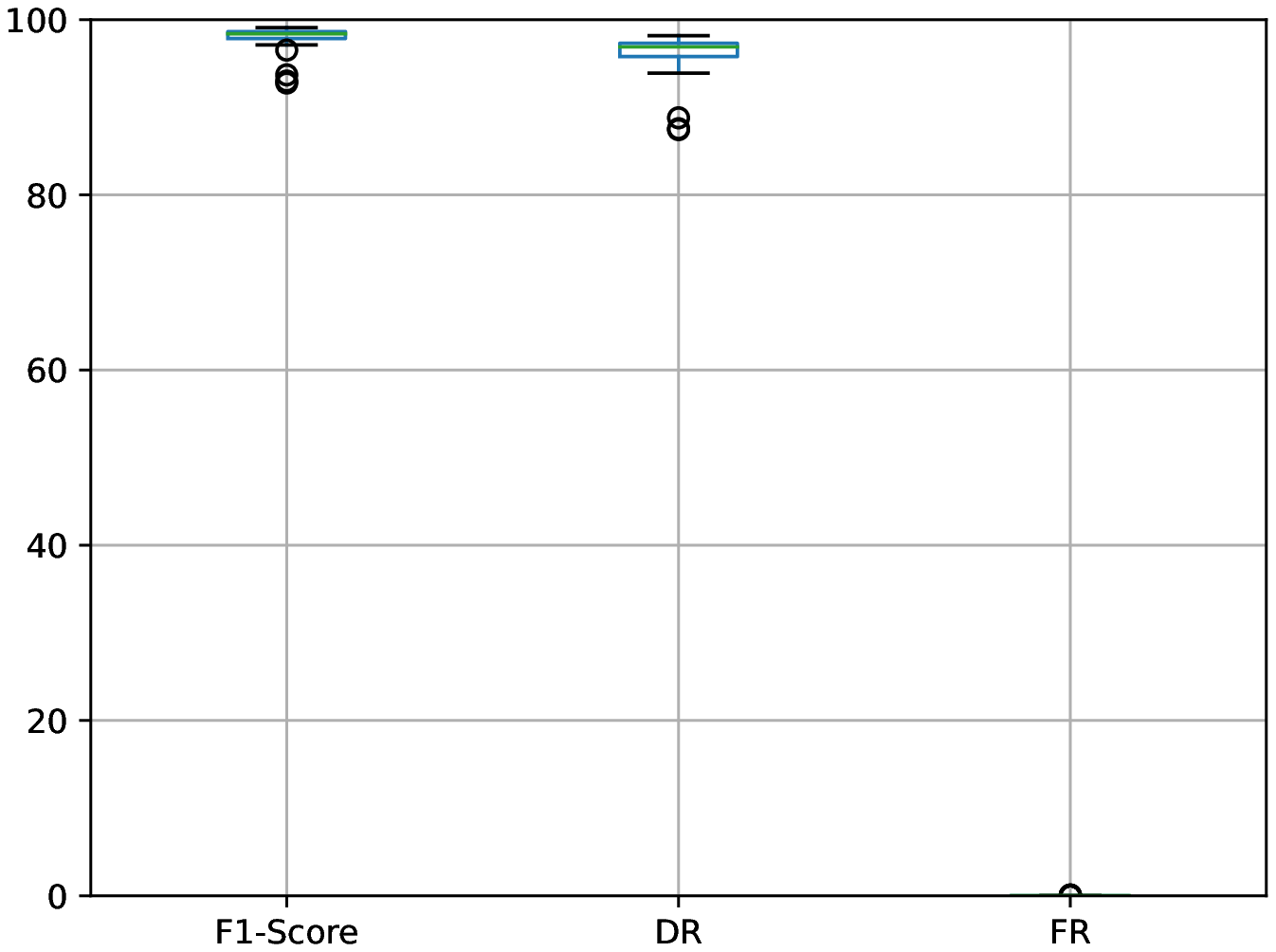}}%
	\qquad
	\subfigure[FedTransformer IEEE-57]{%
		\centering
		\label{fig:box-plot-fedtransformer-57}%
		\includegraphics[width=.32\linewidth]{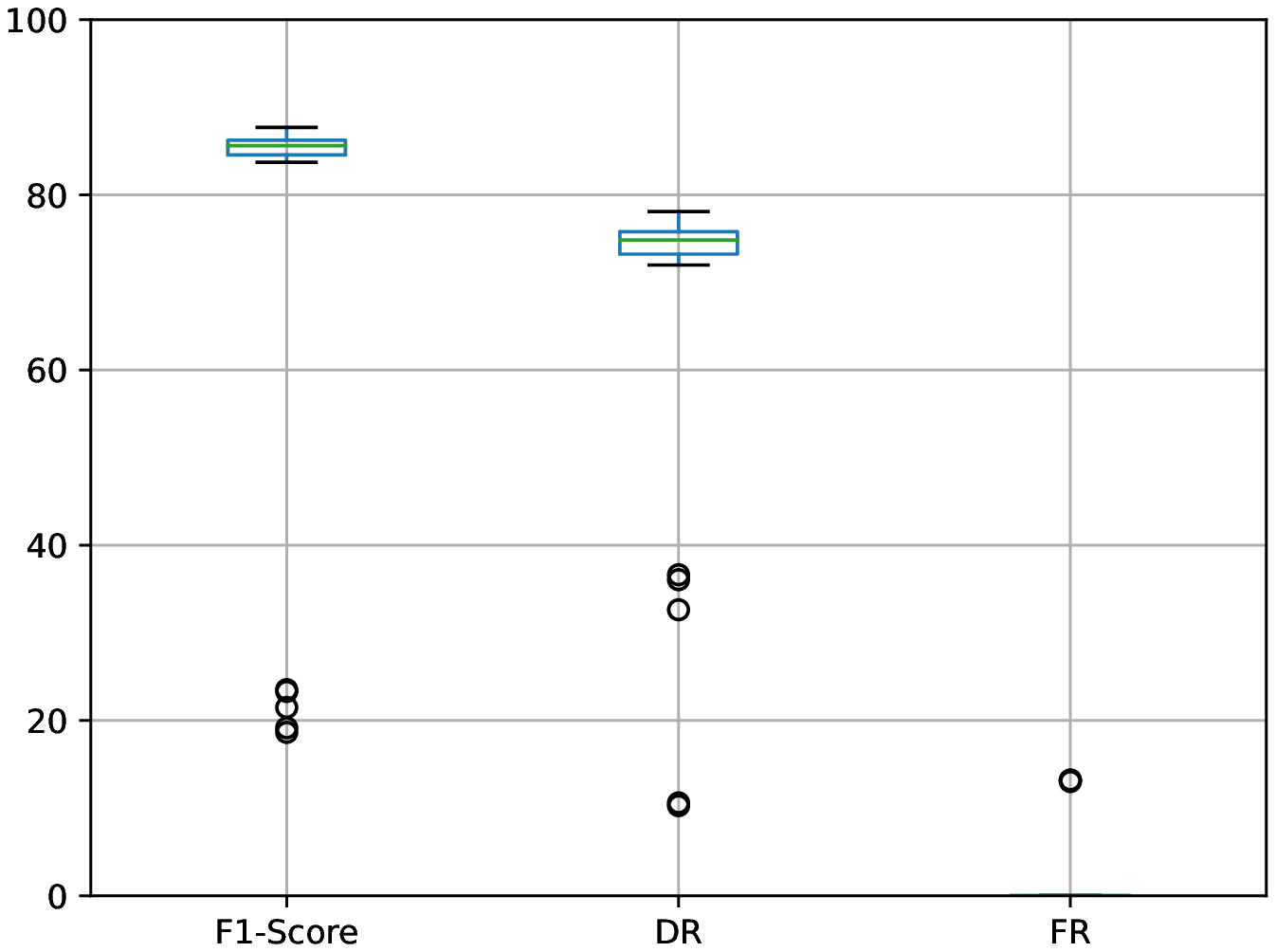}}%
	\qquad
	\subfigure[FedLSTM IEEE-57]{%
		\centering
		\label{fig:box-plot-fedlstm-57}%
		\includegraphics[width=.32\linewidth]{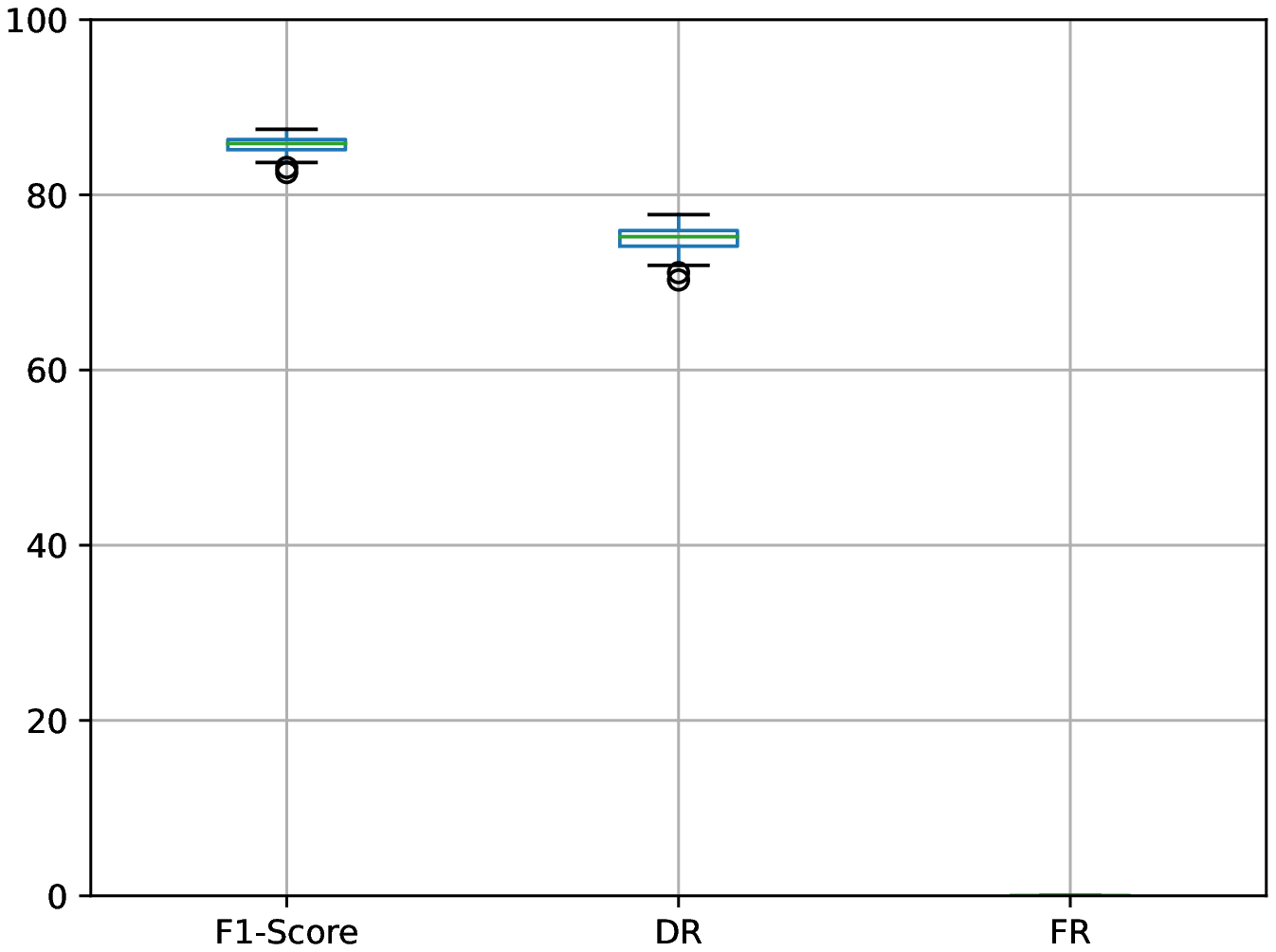}}%
	\qquad
	\subfigure[FedMLP IEEE-57]{%
		\centering
		\label{fig:box-plot-fedmlp-57}%
		\includegraphics[width=.32\linewidth]{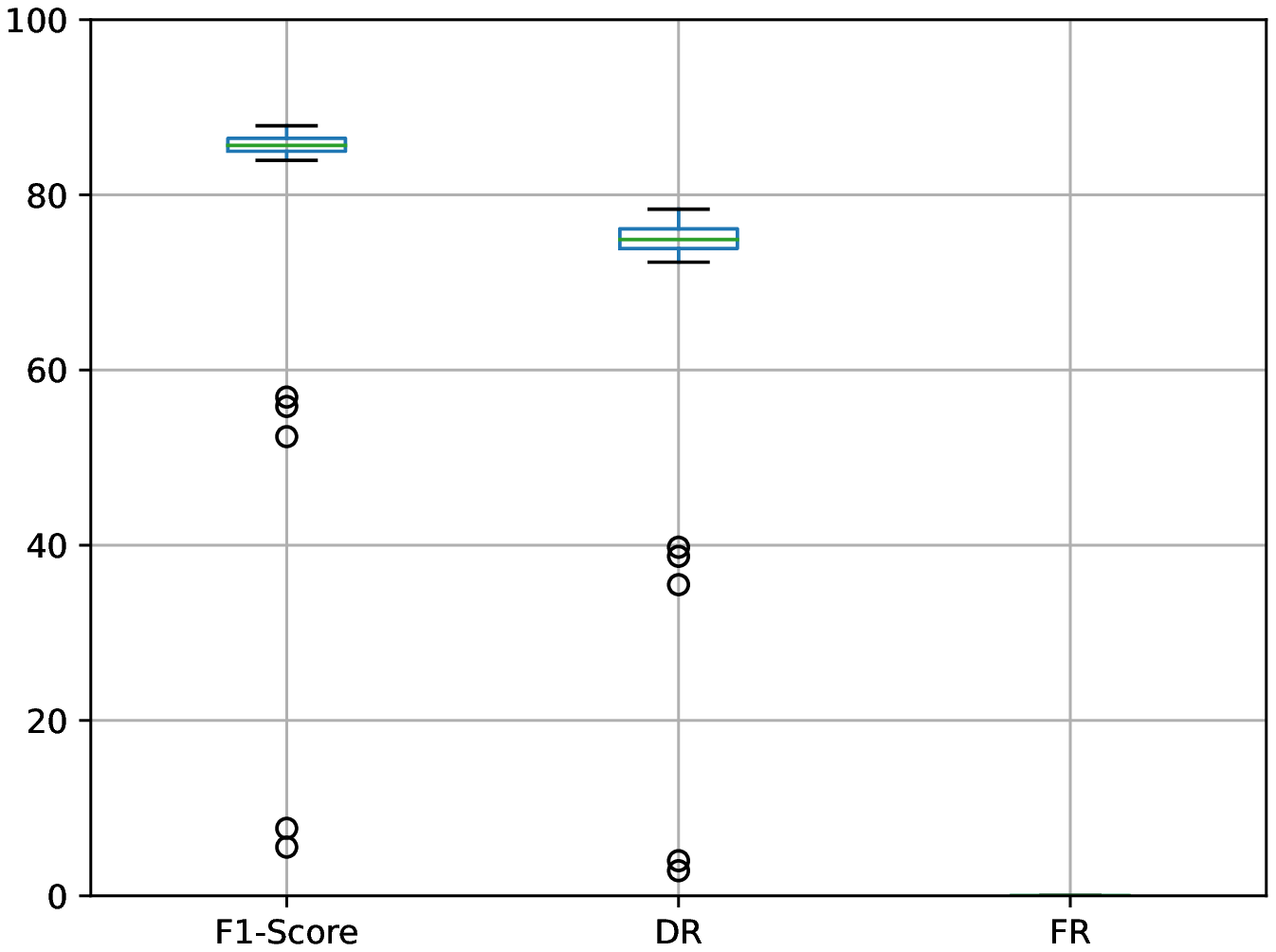}}%
	\caption{Box plots showing the distribution of F1-Score, detection rate, and false alarm rate (in percentage) for each bus in the network for each federated learning model for IEEE 57 bus system.}
	\label{fig:box-plots-57}
\end{figure*}

While the average F1-score, detection rate, and false alarm rate metrics are able to validate the performance of the \ac{FDIA} detection algorithms, we also need to evaluate the performance of the algorithms on each bus in the network. The individual metrics are important to show the fairness and reliability of the algorithm.
In order to visualize the distribution of F1-Score, detection rate, and false alarm rate values among the buses in the network, we plotted box plots for each algorithm in \FIG{fig:box-plots-57} for IEEE 57 bus system. The federated graph learning algorithm outperforms the other algorithms in terms of  the mean value and standard deviation of the F1-Score because it is able to capture both temporal and spatial patterns in the data. Federated transformer and federated \ac{LSTM} algorithms follow the federated graph learning algorithm with almost similar means because they can capture the temporal patterns in the data. The federated \ac{MLP} performs the worst due to the dense connection of the neurons in each layer of the network.

\begin{figure*}
	\centering
	\subfigure[FedGraph IEEE-118]{%
		\centering
		\label{fig:box-plot-fedgraph-118}%
		\includegraphics[width=.32\linewidth]{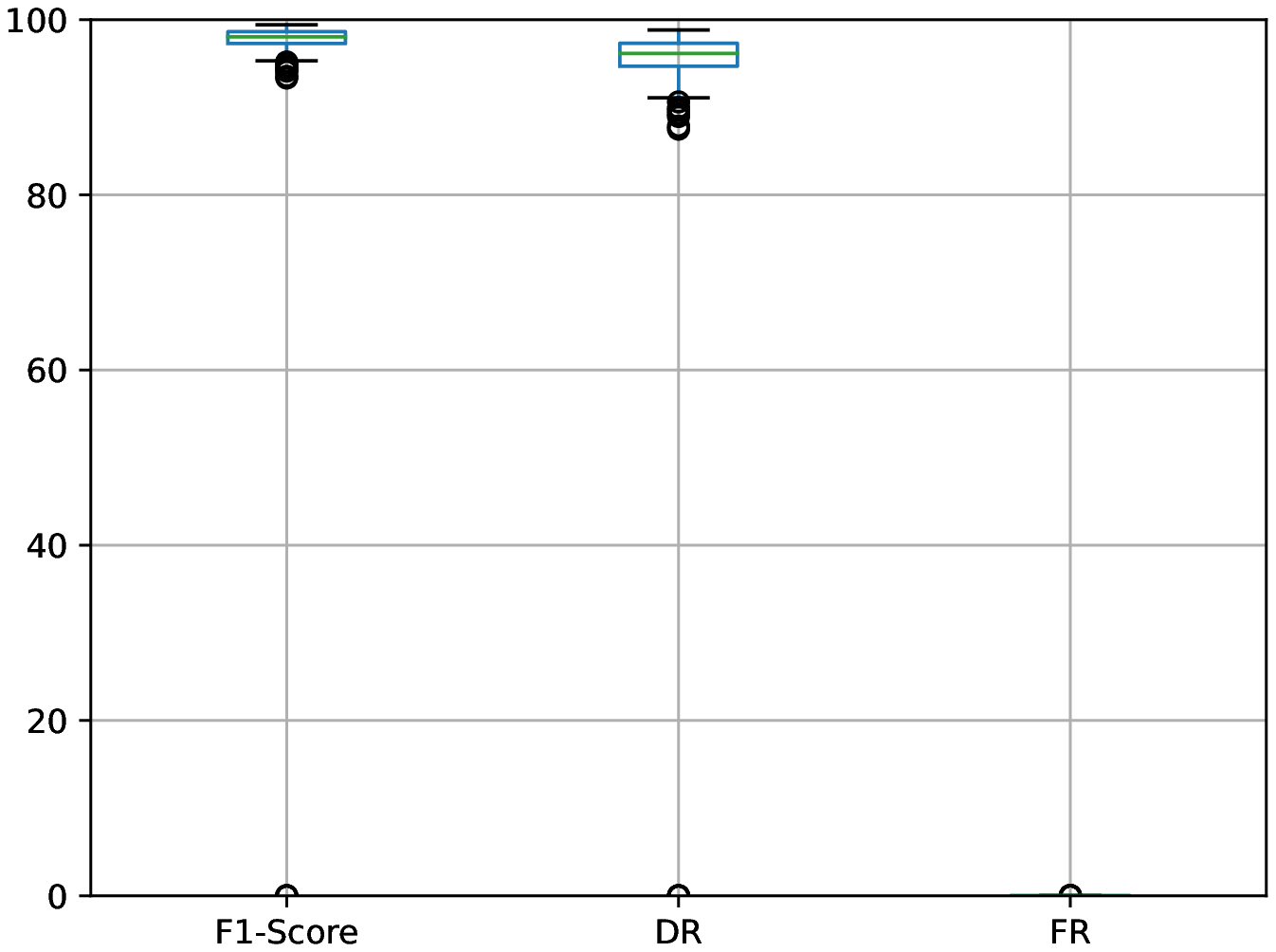}}%
	\qquad
	\subfigure[FedTransformer IEEE-118]{%
		\centering
		\label{fig:box-plot-fedtransformer-118}%
		\includegraphics[width=.32\linewidth]{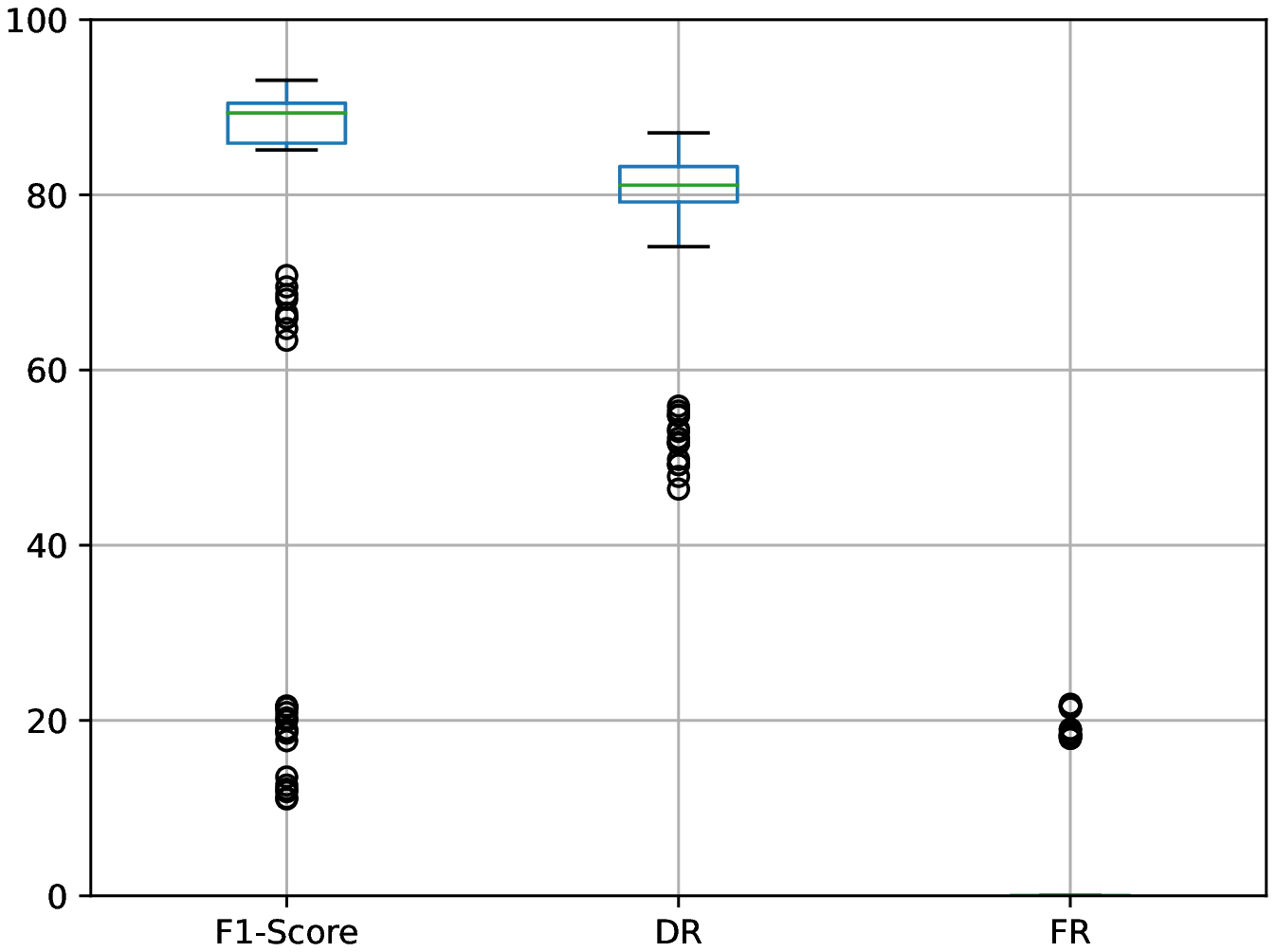}}%
	\qquad
	\subfigure[FedLSTM IEEE-118]{%
		\centering
		\label{fig:box-plot-fedlstm-118}%
		\includegraphics[width=.32\linewidth]{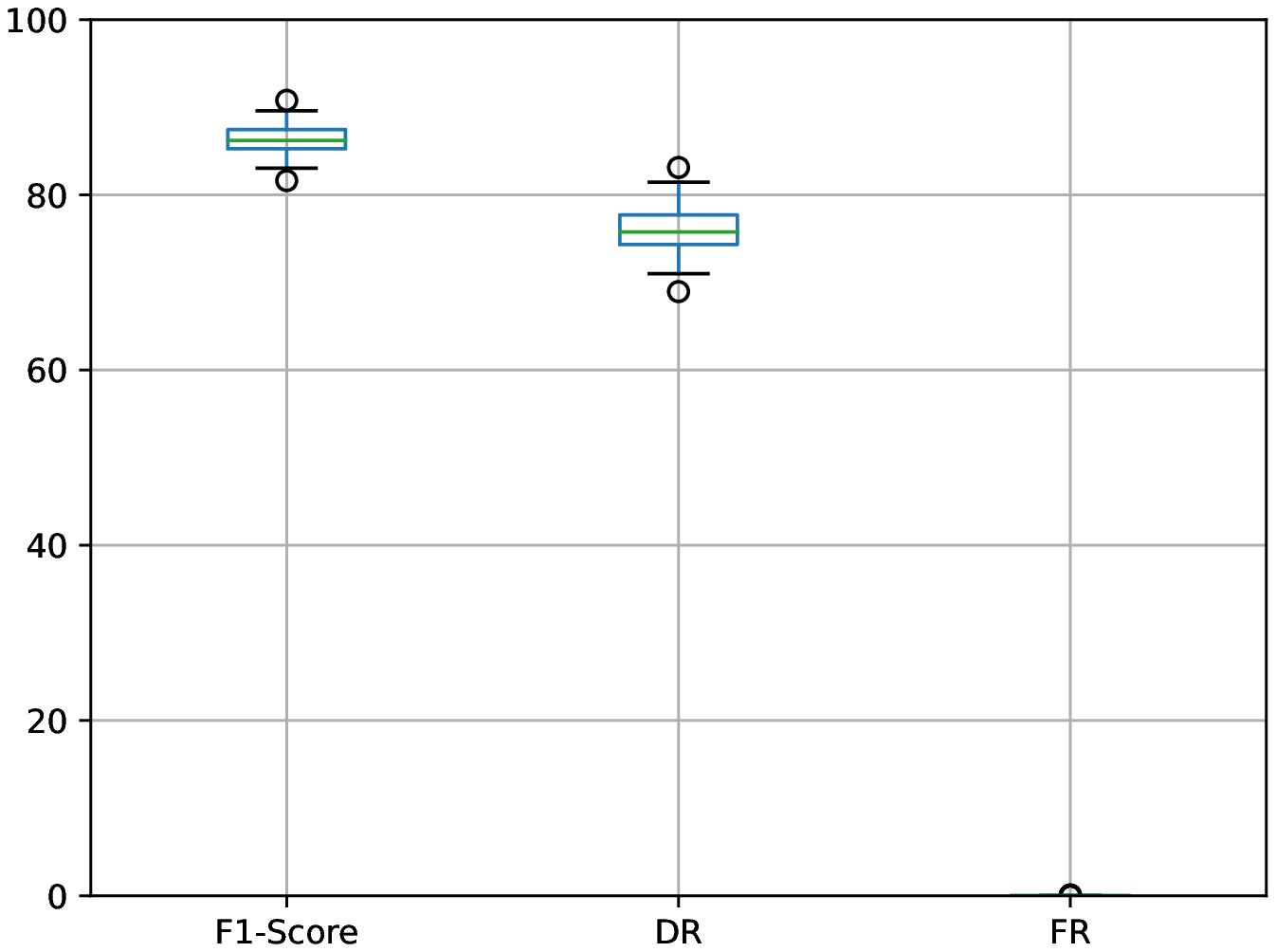}}%
	\qquad
	\subfigure[FedMLP IEEE-118]{%
		\centering
		\label{fig:box-plot-fedmlp-118}%
		\includegraphics[width=.32\linewidth]{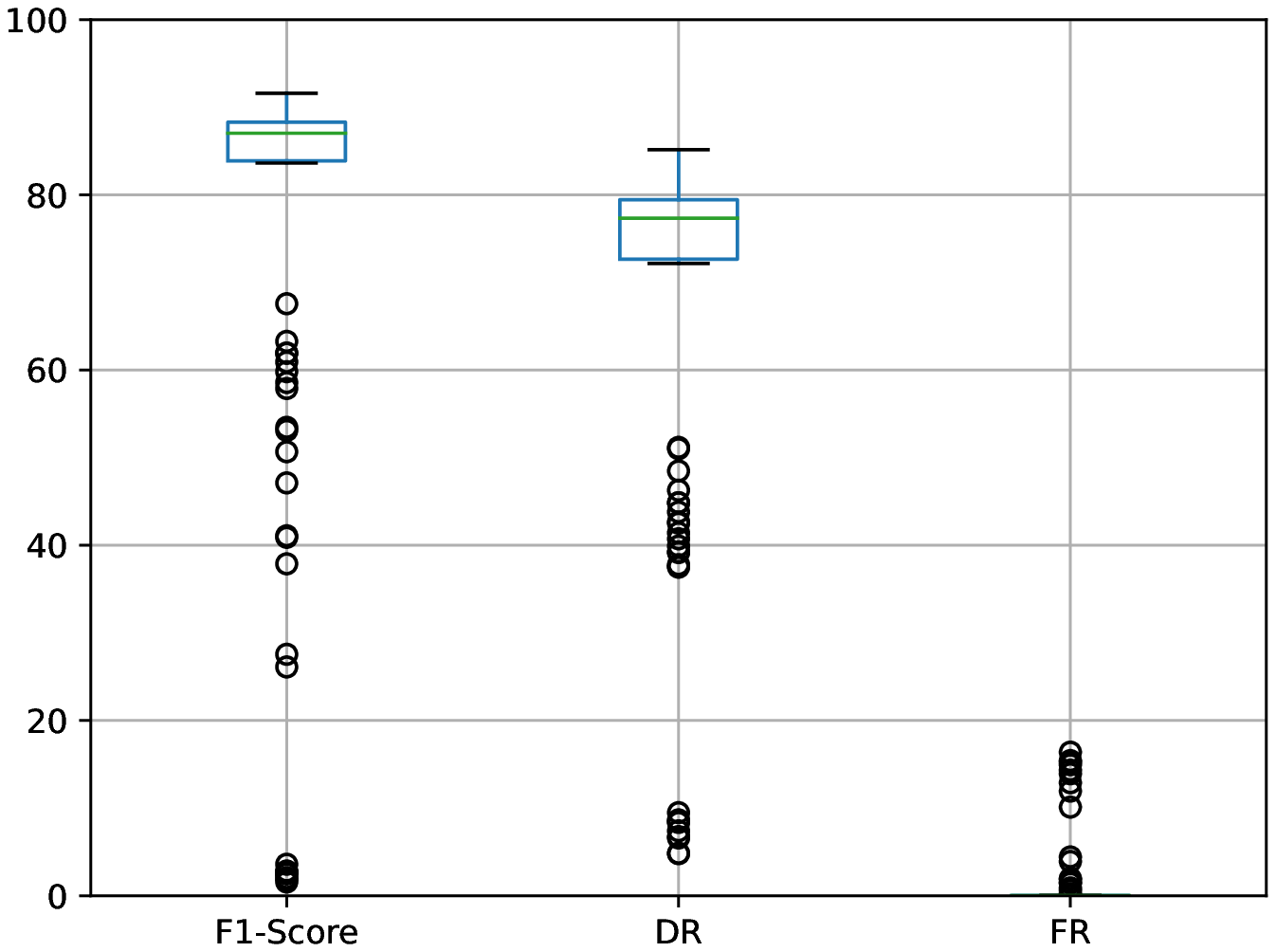}}%
	\caption{Box plots showing the distribution of F1-Score, detection rate, and false alarm rate for each bus in the network for each federated learning model for IEEE 118 bus system.}
	\label{fig:box-plots-118}
\end{figure*}

\begin{figure*}
	\centering
	\subfigure[FedGraph IEEE-300]{%
		\centering
		\label{fig:box-plot-fedgraph-300}%
		\includegraphics[width=.32\linewidth]{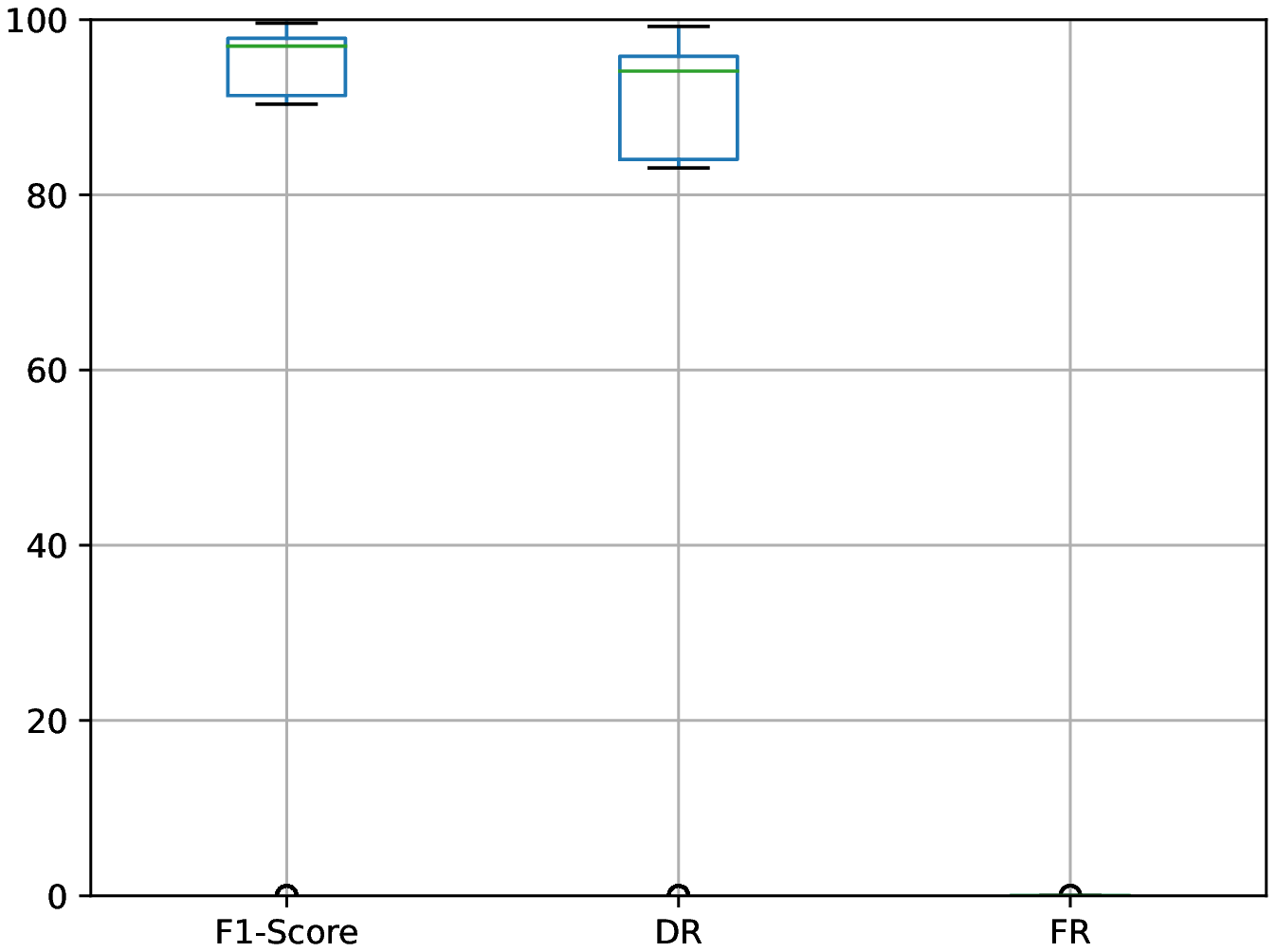}}%
	\qquad
	\subfigure[FedTransformer IEEE-300]{%
		\centering
		\label{fig:box-plot-fedtransformer-300}%
		\includegraphics[width=.32\linewidth]{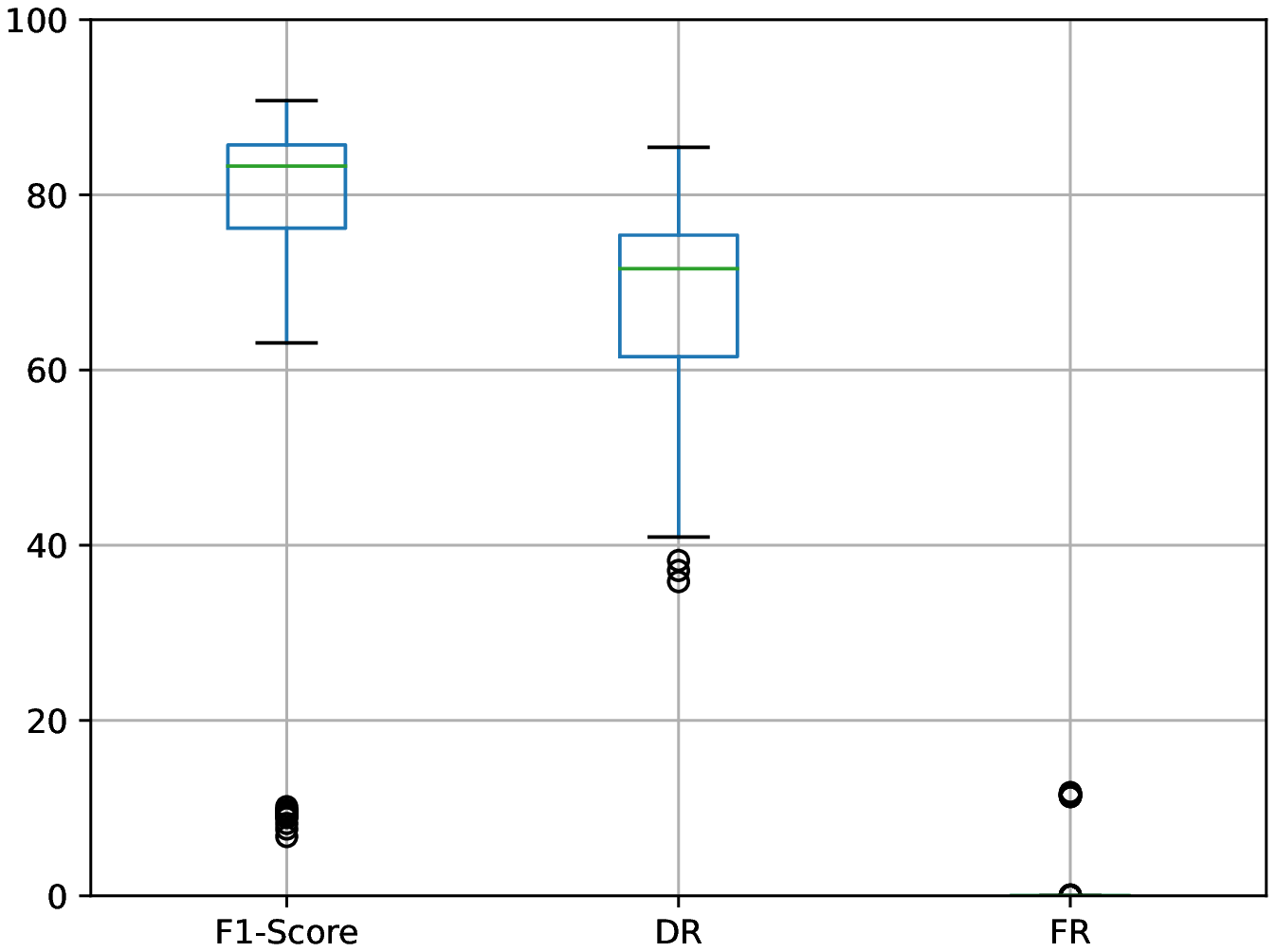}}%
	\qquad
	\subfigure[FedLSTM IEEE-300]{%
		\centering
		\label{fig:box-plot-fedlstm-300}%
		\includegraphics[width=.32\linewidth]{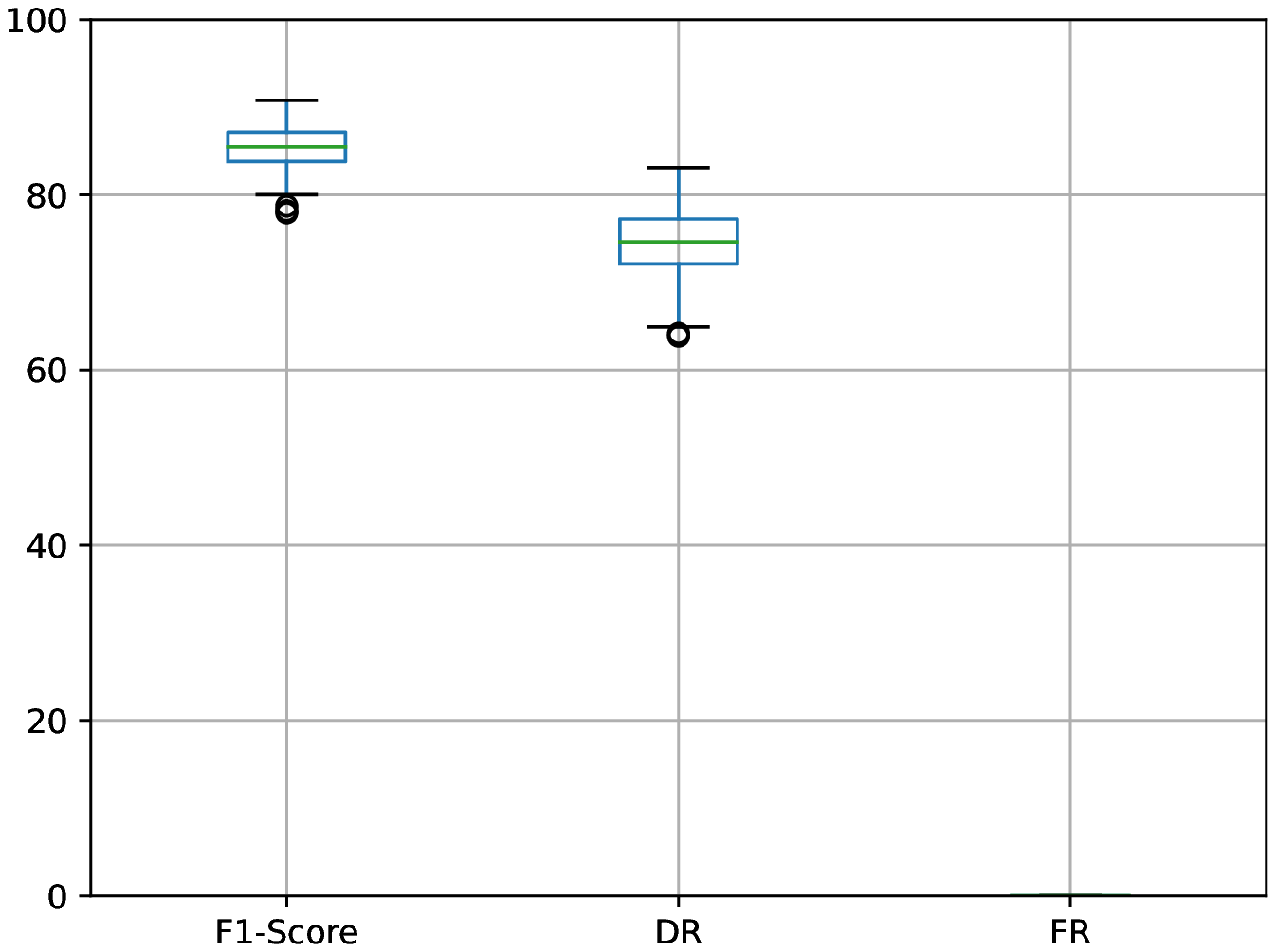}}%
	\qquad
	\subfigure[FedMLP IEEE-300]{%
		\centering
		\label{fig:box-plot-fedmlp-300}%
		\includegraphics[width=.32\linewidth]{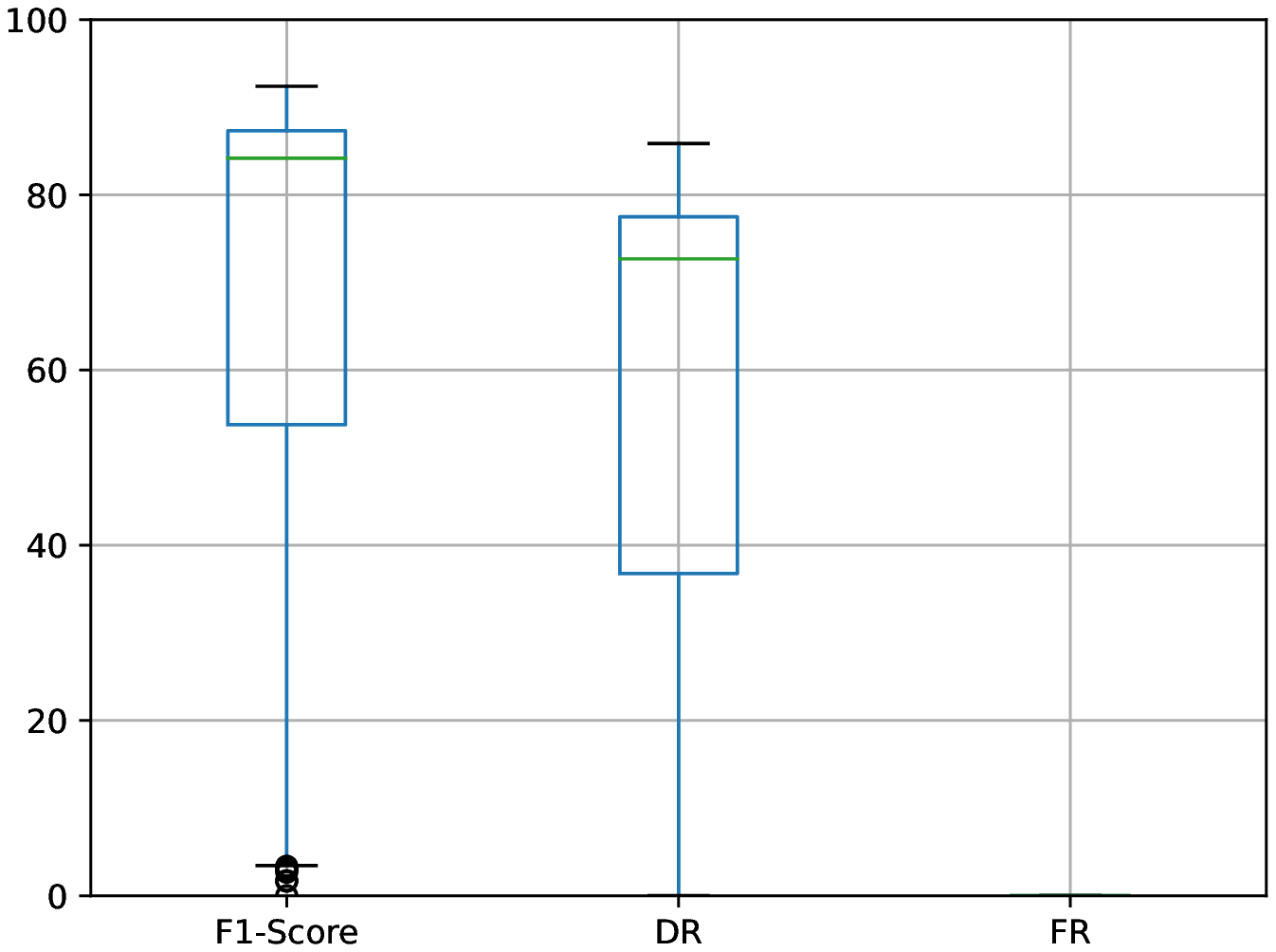}}%
	\caption{Box plots showing the distribution of F1-Score, detection rate, and false alarm rate for each bus in the network for each federated learning model for IEEE 300 bus system.}
	\label{fig:box-plots-300}
\end{figure*}

The box plots showing the detection metrics for the IEEE 118 bus system are provided in Fig. \ref{fig:box-plots-118}. The proposed architecture outperforms the other algorithms, again, for all the metrics. While the federated transformer performs better than the \ac{LSTM} architecture in the mean values of the metrics, it has more outliers and provides less fairness among the clients. Federated \ac{MLP} algorithm performs the worst in terms of the mean and  standard values of the detection metrics. The box plots for the detection metrics for IEEE 300 bus system are displayed in Fig. \ref{fig:box-plots-300}. The results are similar to the ones reported for the IEEE 118 bus system.

\section{Conclusion}
\label{sec:conclusion}
This paper  proposed a hybrid neural network architecture consisting of \ac{LSTM} and \ac{GCN} layers for federated learning based \ac{FDIA} detection. The proposed architecture is able to exploit the temporal and spatial patterns in the data in order to efficiently detect the \ac{FDIA} in smart grids. We proposed to use the conventional \ac{FedAvg} algorithm for aggregating the feature extractor weights of each client while using the FedGraph algorithm to aggregate the weights of the \ac{GCN} layers. This scheme enables efficient and robust training of the deep learning-based detection method. The use of \ac{GCN} layers combined with the FedGraph algorithm allows federated training on any partition of the power networks unlike the existing algorithms in the literature, which can be used only at the node level. We performed extensive experiments on each of the IEEE 57, 118, and 300 bus systems combined with real power usage data. The experiment results show that the proposed algorithm outperforms the state-of-the-art algorithms.

\bibliography{main.bib}
\bibliographystyle{IEEEtran}

\end{document}